\documentclass[10pt,twocolumn,letterpaper]{article}

\usepackage[pagenumbers]{cvpr} %

\usepackage{graphicx}
\usepackage{amsmath}
\usepackage{amssymb}
\usepackage{booktabs}
\usepackage[accsupp]{axessibility}

\usepackage[pagebackref,breaklinks,colorlinks]{hyperref}

\usepackage[capitalize]{cleveref}
\crefname{section}{Sec.}{Secs.}
\Crefname{section}{Section}{Sections}
\Crefname{table}{Table}{Tables}
\crefname{table}{Tab.}{Tabs.}

\def\cvprPaperID{3407} %

\def\confName{CVPR}
\def\confYear{2023}

\usepackage{bm}
\usepackage{booktabs}
\usepackage{caption}
\usepackage{color}
\usepackage{enumitem}
\usepackage{multirow}
\usepackage{url}

\newcommand{\std}[1]{\tiny $\pm$#1}

\usepackage[capitalize]{cleveref}
\crefname{section}{Sec.}{Secs.}
\Crefname{section}{Section}{Sections}
\Crefname{table}{Table}{Tables}
\crefname{table}{Tab.}{Tabs.}

\begin{document}

\title{Towards Flexible Multi-modal Document Models}

\author{Naoto~Inoue$^{1}$ \qquad Kotaro~Kikuchi$^{1}$ \qquad Edgar~Simo-Serra$^{2}$ \qquad Mayu~Otani$^{1}$ \qquad Kota~Yamaguchi$^{1}$\\
$^{1}$CyberAgent, Japan \qquad $^{2}$Waseda University, Japan\\
{\tt\small \{inoue\_naoto,~kikuchi\_kotaro\_xa\}@cyberagent.co.jp~ess@waseda.jp} \\
{\tt\small \{otani\_mayu,~yamaguchi\_kota\}@cyberagent.co.jp}
}

\maketitle
\thispagestyle{empty}

\begin{abstract}
Creative workflows for generating graphical documents involve complex inter-related tasks, such as aligning elements, choosing appropriate fonts, or employing aesthetically harmonious colors.
In this work, we attempt at building a holistic model that can jointly solve many different
design tasks.
Our model, which we denote by \emph{FlexDM}, treats vector graphic documents as a set of multi-modal elements, and learns to predict masked fields such as element type, position, styling attributes, image, or text, using a unified architecture.
Through the use of explicit multi-task learning and in-domain pre-training, our model can better capture the multi-modal relationships among the different document fields.
Experimental results corroborate that our single FlexDM is able to successfully solve a multitude of different design tasks, while achieving performance that is competitive with task-specific and costly baselines.
\footnote{Please find the code and models at: \newline \url{https://cyberagentailab.github.io/flex-dm}.}
\end{abstract}

\section{Introduction}
\label{sec:intro}

Vector graphic documents are composed of diverse multi-modal elements such as
text or images and serve as the dominant medium for visual communication today.
The graphical documents are created through many different design tasks,
\eg, filling in a background image, changing font and color, adding a
decoration, or aligning texts.  While skilled designers perform tasks based on
their design knowledge and expertise, novice designers often struggle to make
decisions to create an effective visual presentation.  To assist such novice
designers, interactive frameworks equipped based on models
that learn design knowledge from completed
designs have been proposed~\cite{odonovan2015designscape,guo2021vinci}.  Our present work
proposes models that can be used in such systems, with a particular focus on
developing holistic models that can flexibly switch between design tasks.

Design tasks are characterized by 1) the variety of possible actions and 2) the
complex interaction between multi-modal elements.
As discussed above, a designer can make almost any edit to the appearance of a
vector graphic document, ranging from basic layout to nuanced font
styling.
While there have been several studies in solving specific tasks of a single
modality, such as layout generation~\cite{li2019layoutgan,gupta2021layout,arroyo2021variational,kong2022blt,kikuchi2021modeling}, font recommendation~\cite{zhao2018modeling}, or colorization~\cite{kikuchi2023generative,yuan2021infocolorizer,qiu2023color}, in realistic design
applications, we believe it is essential to build a \emph{flexible} model that
can consider multiple design tasks in a principled manner to make automated
decisions on creative workflow.

\begin{figure*}[t]
  \centering
  \includegraphics[width=\linewidth]{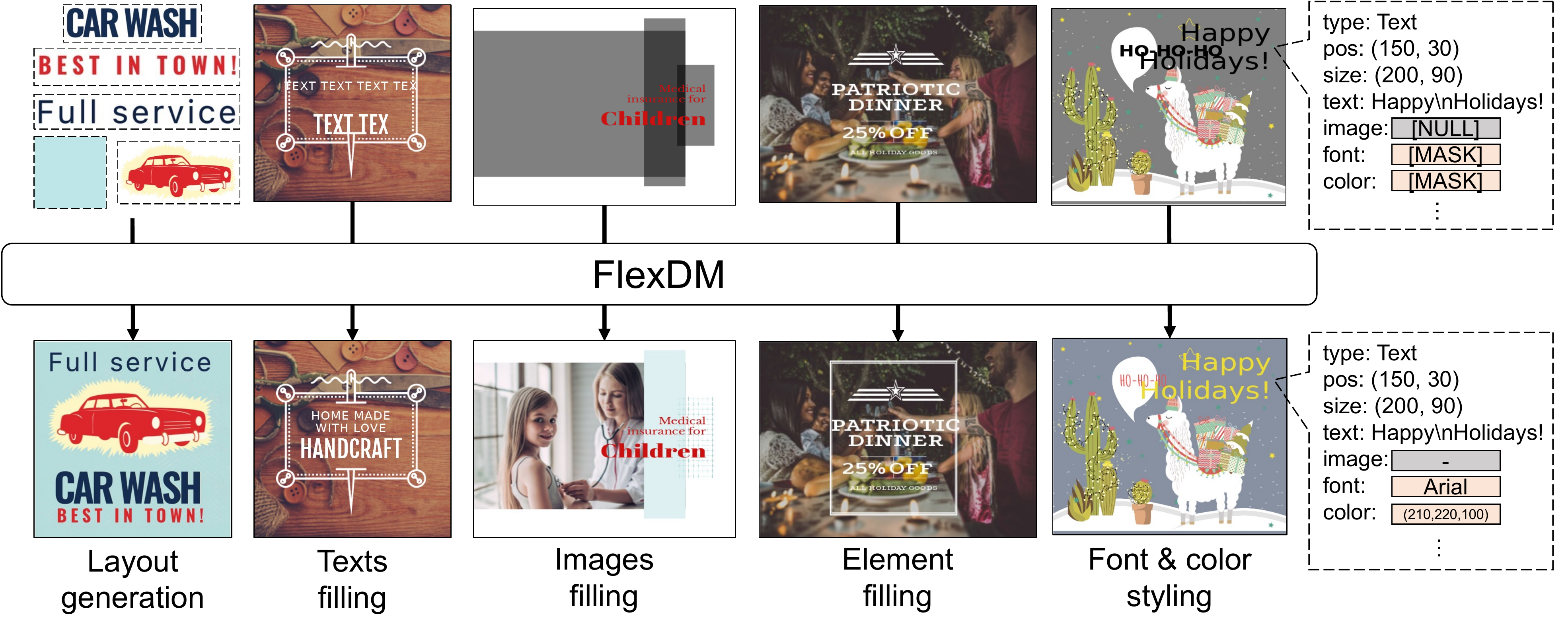}
  \caption{
    Examples of the design tasks that can be solved by our proposed FlexDM
    model, which is designed to process a vector graphic document consisting of
    an arbitrary number of elements (\eg, text).  Each element is composed of
    multi-modal fields indicating its attribute properties (\eg, text content,
    position, font color, etc.).
  }
  \label{fig:teaser}
\end{figure*}

In this work, we refer to a certain attribute of an element as a \emph{field}
and formulate the various design tasks as a unified \emph{masked field
prediction}, which is inspired by the recent masked
autoencoders~\cite{devlin-etal-2019-bert,he2022masked} and multi-task
models~\cite{jaegle2022perceiverio,lu2023unifiedio}.
The key idea is to utilize masking patterns to switch among
different design tasks within a single model; \eg, element filling can be
formulated as predicting all the fields of the newly added element.
Our flexible document model, denoted by \emph{FlexDM}, consists of an
encoder-decoder architecture with a multi-modal head dedicated to handling
different fields within a visual element.  After pre-training with random
masking strategy, we train FlexDM by explicit multi-task learning where we
randomly sample tasks in the form of masking patterns corresponding to the
target design task.
We illustrate in \cref{fig:teaser,fig:overview_concepts} an overview of FlexDM,
with emphasis on the correspondence between design tasks and masking patterns.

Through our carefully designed experiments, we show that our proposed FlexDM performs favorably
against baselines in five design tasks using the Rico~\cite{deka2017rico} and
Crello~\cite{yamaguchi2021canvasvae} datasets.
We also study how different modeling approaches affect the final task
performance in the ablation study.
Finally, we apply our framework to several previously studied design tasks with
minimal modifications and show that the performance matches or even surpasses
the current task-specific approaches.

Our contributions can be summarized in the following.
\begin{itemize}[noitemsep,nolistsep,leftmargin=*]
  \item We formulate multiple design tasks for vector graphic documents by masked multi-modal field prediction in a set of visual elements.
  \item We build a flexible model to solve various design tasks jointly in a single Transformer-based model via multi-task learning.
  \item We empirically demonstrate that our model constitutes a strong baseline for various design tasks.
\end{itemize}

\section{Related Work}
\label{sec:related_work}

\subsection{Vector Graphic Generation}
\label{subsec:rw_vg_generation}
There has been a growing interest in vector graphics to realize resolution/artifact-free rendering that is easy to interpret and edit, such as Scalable Vector Graphics (SVG)~\cite{svg}.
Modeling documents in a vector format is much more complex than the stroke or path level vector graphics~\cite{ha2017neural,lopes2019learned,carlier2020deepsvg} since each element contains multi-modal features such as text and image.
CanvasVAE~\cite{yamaguchi2021canvasvae} tackles the document-level unconditional generation of vector graphics, but is not a multi-task model and cannot solve specific design tasks such as element filling.
Doc2PPT~\cite{fu2022doc2ppt} generates slides given a longer and more detailed multi-modal document, but it is a summarization task and cannot infer what is missing in the incomplete document.

Obtaining transferable representation for downstream tasks learned from multi-modal large-scale data is getting popular.
Domains closest to our setting are document understanding~\cite{xu2020layoutlm,xu2020layoutlmv2,li2021selfdoc,xie2021canvasemb} and UI understanding~\cite{he2021actionbert,chongyang2021uibert}, where the data consist of elements with multi-modal attributes.
Despite the generalizable representation, all the methods fine-tune different parameters for each downstream task (mainly in classification).
In contrast, we aim to solve many essential tasks for design creation in a single model.

\subsection{Multi-task Learning}
\label{subsec:rw_mtl}
Multi-task learning (MTL)~\cite{caruana1997multitask,evgeniou2004regularized,argyriou2006multi} aims at solving different tasks at the same time while sharing information and computation among them, which is crucial for deployment.
MTL methods achieve a good tradeoff between performance and computational cost by (i) multiple lightweight heads at the top of shared backbone~\cite{zhang2014facial,kokkinos2017ubernet} and (ii) efficient use of task-specific parameters~\cite{rebuffi2017learning,rebuffi2018efficient,liu2019end}.
On the contrary, our model obtains the task information from the masking patterns of the input fields and we empirically show that extra task-specific parameters are not necessary.

Training a single model that generalizes to many different tasks has been a long-standing goal.
12-in-1~\cite{lu202012} and UniT~\cite{Hu_2021_ICCV} handle multiple tasks in vision and language domain with small task-specific parameters.
In a more unified manner,
Perceiver~\cite{jaegle2021perceiver} and Perceiver IO~\cite{jaegle2022perceiverio} treat different modalities as the same data format, OFA~\cite{wang2022ofa} and Unified-IO~\cite{lu2023unifiedio} consider similar attempts in the sequence-to-sequence framework,
resulting in a single model or architecture with no task-specific tuning.
We are highly inspired by these works and explore how to unify the design tasks in vector graphic document domain.

\subsection{Computational Assistance for Graphic Design}
\label{subsec:rw_computational_assistance}
There is a long history of automatic graphic design~\cite{lok2001survey,agrawala2011design,yang2016automatic}.
Recent approaches rely on the learning-based formulation, where the primal focus is in predicting layouts given label sets~\cite{jyothi2019layoutvae,li2019layoutgan} or in an unconditional manner~\cite{gupta2021layout,arroyo2021variational}, and avoids the manual design of the energy functions seen in the earlier work~\cite{odonovan2014learning}.
Some works additionally take positional/relational constraints~\cite{li2020attribute,lee2020neural,Kikuchi2021} or textual descriptions~\cite{zheng2019content} for finer design control, but are not applicable in a more complex scenario.
In contrast, our multi-task approach solves many conditional tasks thanks to the flexible multi-modal fields in both inputs and targets.

Considering multi-modal features is essential to go beyond layout generation for intelligent graphic design assistance.
Wang~\etal~\cite{wang2020learning} retrieve images from layout information and keywords for each element to obtain visually pleasing visual design by reinforcement learning.
Zhao~\etal~\cite{zhao2018modeling} predict font properties of a text on a webpage over a background image considering metadata.
Li~\etal~\cite{li2021harmonious} predict position and size for a single text box over a background image considering saliency.
We demonstrate that we can apply our flexible model to solve these tasks with almost no modification, and our model performs favorably against the task-specific well-tuned approaches.

\begin{figure}[t]
  \centering
    \includegraphics[width=\hsize]{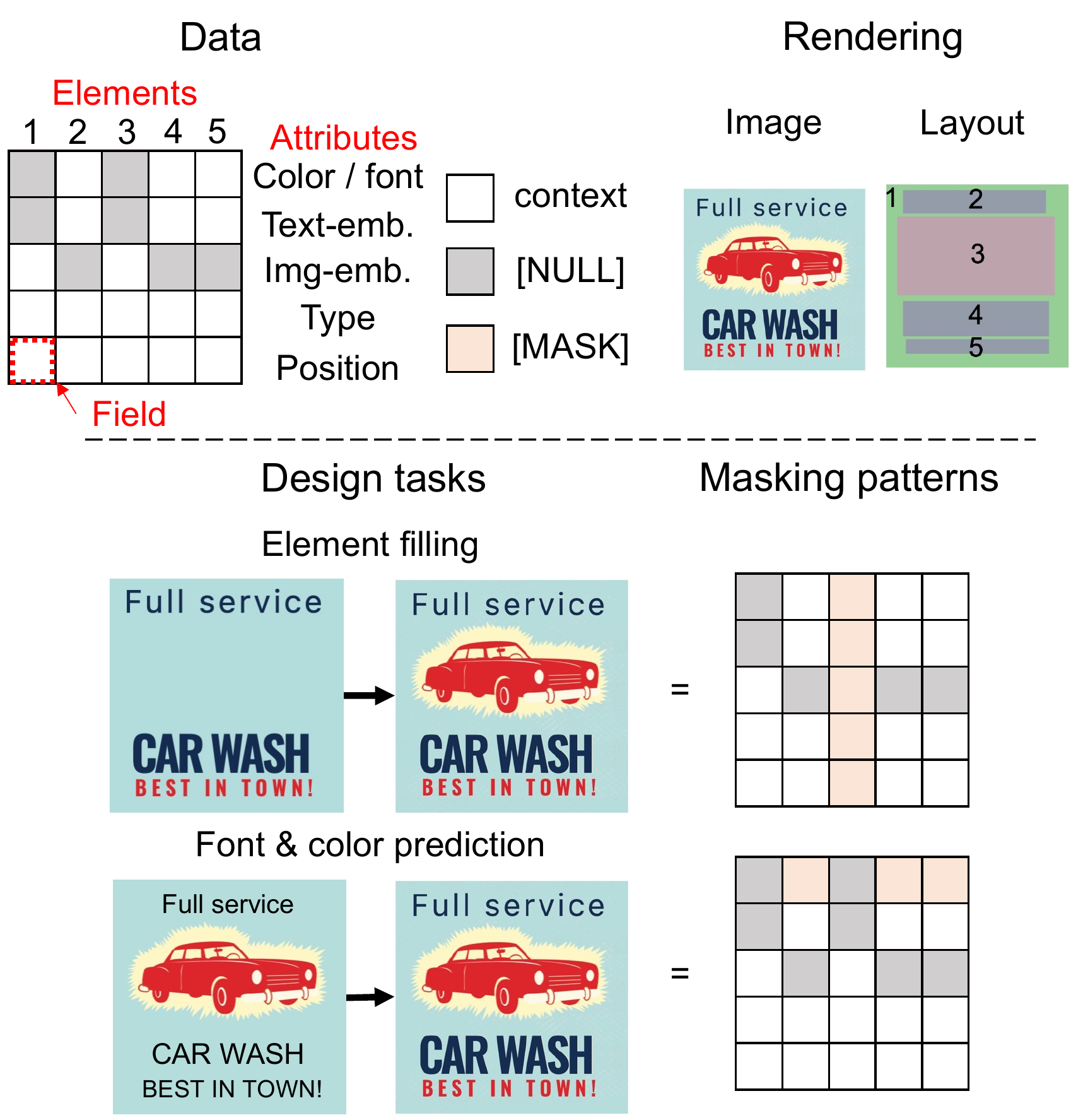}
  \caption{
    \textbf{Top}: example of a vector graphic document consisting of five elements.
    The array is used to illustrate the data structure of the document.
    Each column corresponds to a single visual element. Each row corresponds to an attribute or a group of attributes consisting the element.
    \textbf{Bottom}:
    Correspondence between design tasks and masking patterns for our masked field prediction.
  }
  \label{fig:overview_concepts}
\end{figure}

\begin{figure}[t]
  \centering
  \includegraphics[width=.95\linewidth]{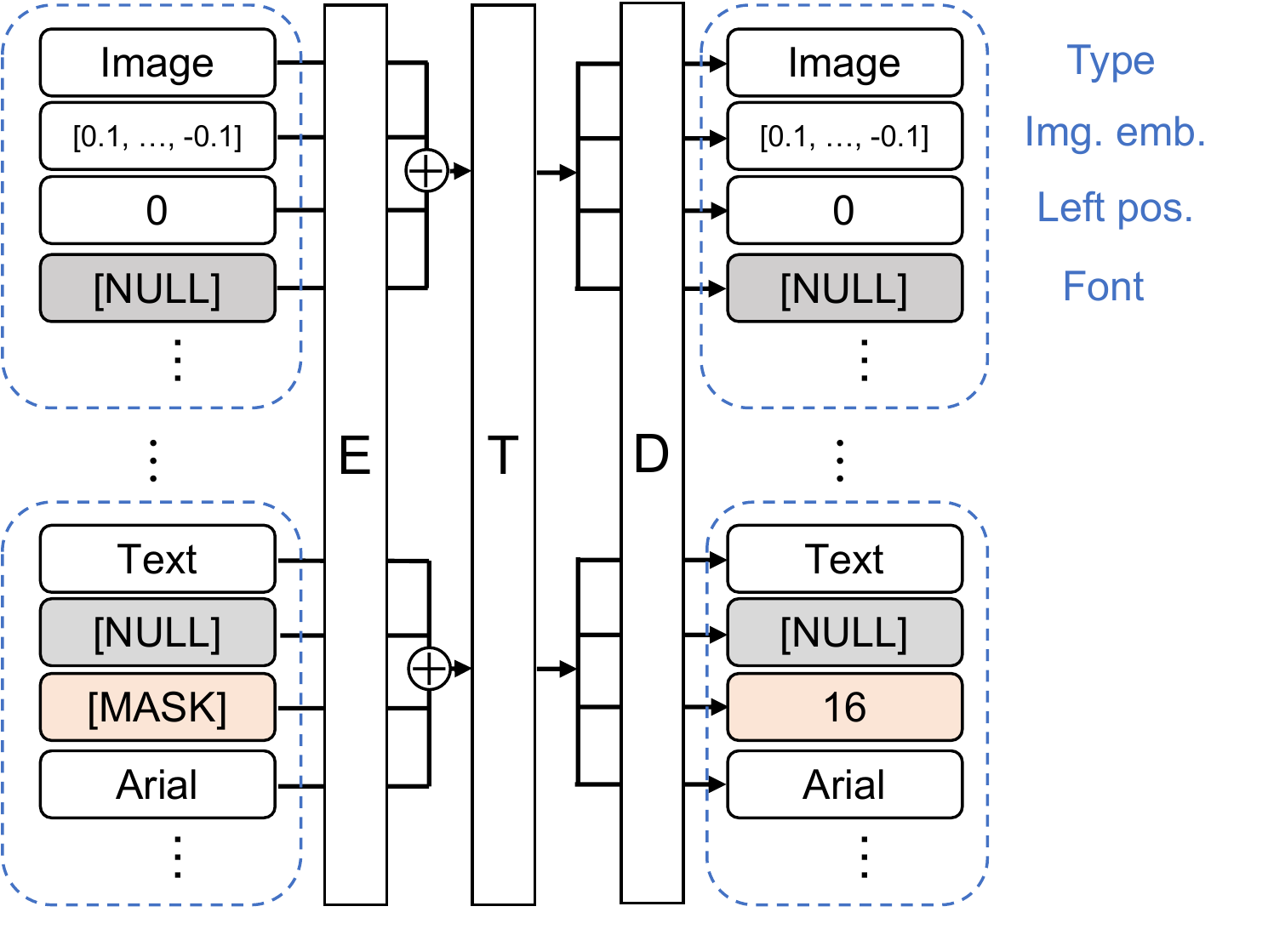}
  \caption{
    The architecture of FlexDM. E, T, and D are short for Encoder, Transformer blocks, and Decoder, respectively.
  }
  \label{fig:unid_architecture}
\end{figure}

\section{Approach}
\label{sec:approach}
We first describe the formal definition of the vector graphic document and
notations in \cref{subsec:preliminary}.
We then introduce the idea of masked field prediction and a model for it in
\cref{subsec:masked_field_prediction} and \cref{subsec:arch}.
Finally, we describe how we train FlexDM in~\cref{subsec:universal_model}.

\subsection{Preliminary}
\label{subsec:preliminary}
\noindent \textbf{Document Structure}:
In this work, a vector graphic document $X$ consists of a set of elements
$X=(X_{1},X_{2}, \ldots, X_{S})$, where $S$ is the number of elements in $X$.
Each element $X_{i}$ consists of a set of multi-modal fields and denoted by
$X_{i}=\{x_{i}^{k} \;|\; k \in \mathcal{E}\}$, where $\mathcal{E}$ indicates
the indices for all the attributes.
Each field $x_{i}^{k}$ can be either a categorical or numerical variable
such as element type, position, text content, or image embedding.
For ease of explanation, we illustrate $X$ by a 2D-array as shown in the top of \cref{fig:overview_concepts}.
Note that the order in the array does not matter because $X$ is a set of sets.
Since processing high-dimensional data such as raw images and texts during optimization is
computationally intensive, we extract a low-dimensional numerical vector
from such data for $x_{i}^{k}$ using pre-trained models.
\\ \noindent \textbf{Special Tokens}:
In a similar spirit to the masked language model~\cite{devlin-etal-2019-bert},
we use a few special tokens to represent $x_{i}^{k}$.\\
$\texttt{[NULL]}$: appears when $x_{i}^{k}$ is inevitably missing (\eg, font type
for an image element), or padding variable-length sequence within a mini-batch on training. \\
$\texttt{[MASK]}$: appears when $x_{i}^{k}$ is masked for prediction.

\subsection{Masked Field Prediction}
\label{subsec:masked_field_prediction}
Given an incomplete document $X$ containing $\texttt{[MASK]}$ as context, our
goal is to predict values for all the fields filled with $\texttt{[MASK]}$ and
generate a complete document $\hat{X}$.
We refer to this problem by \textit{masked field prediction}, where a model has to
predict the masked field considering the different multi-modal relations between
the fields.
While the masking approach is similar to the masked language
model~\cite{devlin-etal-2019-bert}, there is a key distinction in that we
process an order-less set of multi-modal items (\ie, document $X$).
For this reason, we design our architecture to 1) efficiently capture inter-field
relationships of vector graphic attributes, and 2) ensure
that the model works without
positional encodings commonly used to model an ordered sequence.

\subsection{FlexDM Architecture}
\label{subsec:arch}
As shown in \cref{fig:unid_architecture}, our architecture consists of three modules; encoder, Transformer blocks, and decoder.
Given a document, we first project a set of partially masked fields (\eg, position or
font) into embeddings using the encoder, and then feed the output to the intermediate
Transformer blocks.
The final decoder takes the transformed embeddings and projects them back to
the original fields space.
The Transformer blocks only process $S$ embeddings, which is efficient compared to architecture processing $S \times N$ fields with off-the-shelf Transformer~\cite{vaswani2017attention} directly, when there are $N$ attributes.
In the following, let us denote all model parameters by $\theta$.

\noindent \textbf{Encoder}:
The encoder takes a document input $X$ and embeds it into $h^{\text{enc}} =
\{h_{1}^{\text{enc}}, h_{2}^{\text{enc}}, \ldots, h_{S}^{\text{enc}}\}$ with element-wise operations.
The encoder first maps each field $x_{i}^{k}$ to a fixed dimensional vector with
$f^{\text{enc},k}$, and sums up all the fields in the element to produce a latent
vector for the $i$-th element with:
\begin{equation}
  h_{i}^{\text{enc}} = \sum_{k \in \mathcal{E}} f^{\text{enc},k}(x_{i}^{k};\theta),
\end{equation}
\noindent where $f^{\text{enc},k}$ is an embedding function that retrieves learnable
dense embeddings for each category id if $x_{i}^{k}$ is a categorical variable,
or a simple linear projection layer if $x_{i}^{k}$ is a numerical variable.
We treat the special tokens (\ie, $\texttt{[NULL]}$ and $\texttt{[MASK]}$) in
the same manner to the categorical variable.
\\ \noindent \textbf{Transformer Blocks}:
Transformer blocks take $h^{\text{enc}}$ as input and transform it to
$h^{\text{dec}} = \{h_{1}^{\text{dec}}, h_{2}^{\text{dec}}, \ldots, h_{S}^{\text{dec}}\}$.
We stack these intermediate blocks to process complex inter-element relations.
Our model can stack any off-the-shelf Transformer layer to build up the blocks
$f^{\text{trans}}$:
\begin{equation}
  h^{\text{dec}} = f^{\text{trans}}(h^{\text{enc}};\theta)
\end{equation}
\\ \noindent \textbf{Decoder}:
The final decoder takes $h_{\text{dec}}$ and decodes them back into a document
$\hat{X}=(\hat{X}_{1},\hat{X}_{2}, \ldots, \hat{X}_{S})$, where
$\hat{X}_{i}=\{\hat{x}_{i}^{k}\;|\;k \in \mathcal{E}\}$.
We compute each $\hat{x}_{i}^{k}$ by a linear layer $f^{\text{dec},k}$ for both
categorical and numerical variables:
\begin{equation}
  \hat{x}_{i}^{k} = f^{\text{dec},k}(h_{i}^{\text{dec}};\theta).
\end{equation}
\\ \noindent \textbf{Loss}:
We train our model using reconstruction losses.
Let us denote by $X^{*}$ the ground truth of the incomplete document
$X$, and also denote by $M$ a set of tuples indicating the indices for
$\texttt{[MASK]}$ tokens in $X$.
We define the loss function by:

\begin{equation}
  \mathcal{L} = \sum_{(i, k) \in M} l^{k}(\hat{x}_{i}^{k}, {x^{*}}_{i}^{k}), \label{eq:objective}
\end{equation}
\noindent where $l^{k}$ is the loss function for the $k$-th attribute.
For each $l^{k}$, we use softmax cross-entropy loss for categorical
variables and mean squared error for numerical variables.

\subsection{FlexDM Training}
\label{subsec:universal_model}
Masked field prediction allows us to represent diverse design tasks having various
input/output formats just by altering the masking pattern.
The pattern can be both deterministic or stochastic.
The bottom of \cref{fig:overview_concepts} illustrates example tasks and the
corresponding masking patterns.
Although we can formulate arbitrary tasks with masked field prediction, we
consider several subsets of representative design tasks for our evaluation and
analyses in \cref{sec:exp}.

We describe typical masking patterns in the following.
Note that fields already filled with \texttt{[NULL]} will never be replaced in priority to the masking operations.
\emph{Element masking} randomly selects elements and masks all the fields within
the element; \ie, we can formulate the element filling task by single element masking.
\emph{Attribute masking} randomly selects attributes and mask the fields across all the elements; \eg, masking position and size of all the elements becomes
layout prediction, and masking fonts becomes font prediction.
\emph{Random masking} strategy masks fields by some probability without considering
the data structure, which is similar to BERT~\cite{devlin-etal-2019-bert}.

\noindent \textbf{Pre-training}:
To learn the initial model, we employ a pre-training by ordinary random masking similar to the prevailing pre-training strategy of BERT~\cite{devlin-etal-2019-bert}.
One distinction is that our pre-training happens in the same, in-domain dataset, unlike the common setup where a model is pre-trained on a larger dataset in a different domain and then fine-tuned on a target task in a target dataset.
We show in \cref{sec:exp} that this in-domain pre-training moderately improves the final task performance.

\noindent \textbf{Explicit Multi-task Learning}:
The random masking pre-training above is a solid baseline for any task.
Radford~\etal~\cite{radford2019language} hypothesize that this implicit multi-task training leads to the astonishingly strong zero-shot performance of large language models.
However, the random masking strategy actually produces any task with an extraordinarily low probability as the number of attributes and elements increases.
Instead, we employ the \textit{explicit} masking strategy to maximize the performance on all the target tasks.
During training we randomly sample a task from the target tasks, sample a complete document $X^{*}$, and make the triplet ($X$, $X^{*}, M$) by using the masking pattern associated with the task.
We repeat this procedure to build each mini-batch when training FlexDM.

\section{Experiments}
\label{sec:exp}

\subsection{Dataset}
\label{subsubsec:dataset}
We mainly use two datasets containing vector graphic documents,
Rico~\cite{deka2017rico} and Crello~\cite{yamaguchi2021canvasvae}, to evaluate FlexDM.
We basically follow the setting used in~\cite{yamaguchi2021canvasvae}.
Due to memory limitations, we discard documents having more than fifty elements.
Position, size, and color information are discretized in order to enhance the implicit alignment of multiple elements.
We describe the overview of each dataset.
\\ \noindent \textbf{Rico~\cite{deka2017rico}:}
The dataset collects UI designs from mobile apps. We follow previous
works~\cite{li2019layoutgan,lee2020neural} and exclude elements whose labels are not in the most frequent 13 labels.
We divide the dataset into 45,012 / 5,565 / 5,674 examples for train, validation, and test splits.
\\ \noindent \textbf{Crello~\cite{yamaguchi2021canvasvae}:}
The dataset provides design templates from an online design service.
Crello contains various design formats such as social media posts, banner ads, blog headers, or printed posters.
We divide the dataset into 18,738 / 2,313 / 2,271 examples for train, validation, and test splits.
Please refer to the original paper \cite{yamaguchi2021canvasvae} for the definition of each attribute.
For image and text features, we extract 768-dimensional features using CLIP~\cite{radford2021learning}.
We also additionally extract categorical font information (called \texttt{Font}).
We group the attributes into some groups based on their property.
\textbf{TYPE} denotes \texttt{Type} attribute.
\textbf{POS} denotes \texttt{Position} and \texttt{Size} attributes.
\textbf{IMG} denotes \texttt{Image} attribute.
\textbf{TXT} denotes \texttt{Text} attribute.
\textbf{ATTR} denotes attributes not listed above, and these attributes have a large impact on fine-grained appearance.

\subsection{Tasks}
\label{subsubsec:actual_tasks}
We carefully select tasks to evaluate how our model performs in various design tasks.
We select evaluation tasks such that
(i) they are practical,
(ii) they have various combinations of input/output modalities, and
(iii) the masking ratio is modest.
We impose the masking ratio requirement because the extreme masking ratio makes the task too difficult or trivial to solve and makes the baseline comparison impossible.
\\ \noindent \textbf{Element Filling (ELEM):}
This task is to predict a new element that can enhance the document.
We mask all the attributes of a single element in a complete document during training and evaluation.
\\ \noindent \textbf{Attribute Prediction:}
This task is to predict missing attributes at once in the document, which is very challenging.
We apply attribute masking on a complete document to make the masked inputs
during training and evaluation.
We select an attribute group discussed in ~\cref{subsubsec:dataset} and apply the attribute masking for all the attributes in the group.
We consider each group-level prediction task as an individual task.
Note that we do not consider TYPE prediction since it is too trivial and unrealistic.
Therefore, we have two (POS and ATTR) and four (POS, ATTR, IMG, and TXT) attribute prediction tasks for Rico and Crello, respectively.

\subsection{Evaluation Metrics}
\label{subsubsec:evaluation_metrics}
For each task, we quantitatively evaluate the reconstruction performance. The score $\mathcal{S}$ for each document is computed
by:
\begin{equation}
  \mathcal{S} = \frac{1}{|M|} \sum_{(i, k) \in M} s^{k}(\hat{x}_{i}^{k}, {x^{*}}_{i}^{k}), \label{eq:evaluation_metrics}
\end{equation}
\noindent where $s^{k} \in [0,1]$ is a scoring function for $k$-th attribute.
If the attribute is categorical, $s^{k}$ is an indicator function that takes 1 if $\hat{x}_{i}^{k}$ and ${x^{*}}_{i}^{k}$ are identical, otherwise 0.
For image and text features that are the only numerical attributes in our experiments, we use cosine similarity in $[0, 1]$ scale.
\begin{table*}[t]
    \setlength{\tabcolsep}{1pt}
    \begin{minipage}[c]{0.65\linewidth}
    \centering
    \begin{tabular}{lccccccccccc} \toprule
        \multicolumn{1}{r}{Dataset}                   & \multicolumn{4}{c}{Rico~
        \cite{deka2017rico}} &                   & \multicolumn{6}{c}{Crello~\cite{yamaguchi2021canvasvae}}                                                                                                                                     \\ \cmidrule{2-5} \cmidrule{7-12}
        Model                                         & \#par.                   & ELEM              & POS                        & ATTR              &  & \#par. & ELEM              & POS               & ATTR              & IMG               & TXT               \\ \midrule
        Most-frequent                                 & 0.0x                     & 0.461             & 0.213                      & 0.830             &  & 0.0x   & 0.402             & 0.134             & 0.382             & 0.922             & 0.932             \\
        BERT~\cite{devlin-etal-2019-bert}             & 1.0x                     & 0.517             & \underline{0.238}          & 0.847             &  & 1.0x   & \textbf{0.524}    & 0.155             & 0.632             & 0.935             & 0.949             \\
        BART~\cite{lewis2020bart}                     & 1.2x                     & 0.515             & 0.220                      & 0.714             &  & 1.2x   & 0.469             & 0.156             & 0.615             & 0.932             & 0.945             \\
        CVAE~\cite{jyothi2019layoutvae,lee2020neural} & 1.1x                     & 0.511             & 0.214                      & 0.917             &  & 1.0x   & 0.499             & 0.197             & 0.587             & 0.942             & 0.947             \\  %
        CanvasVAE~\cite{yamaguchi2021canvasvae}       & 1.2x                     & 0.437             & 0.192                      & 0.790             &  & 1.2x   & 0.475             & 0.138             & 0.586             & 0.912             & 0.946             \\  %
        Ours-IMP                                      & 1.0x                     & 0.505             & \textbf{0.259}             & 0.923             &  & 1.0x   & 0.483             & 0.197             & 0.607             & 0.945             & 0.949             \\
        Ours-EXP                                      & 1.0x                     & \underline{0.540} & 0.226                      & \underline{0.937} &  & 1.0x   & 0.499             & \underline{0.218} & \underline{0.679} & \underline{0.948} & \underline{0.952} \\
        Ours-EXP-FT                                   & 1.0x                     & \textbf{0.552}    & 0.215                      & \textbf{0.945}    &  & 1.0x   & \underline{0.508} & \textbf{0.227}    & \textbf{0.688}    & \textbf{0.950}    & \textbf{0.954}    \\ \midrule
        Expert                                        & 3.0x                     & 0.575             & 0.228                      & 0.952             &  & 5.0x   & 0.534             & 0.255             & 0.703             & 0.948             & 0.955             \\
        \bottomrule
    \end{tabular}
    \caption{
        Quantitative evaluation in two datasets.
        A higher score indicates the better performance.
        Top two results are highlighted in \textbf{bold} and \underline{underline}, respectively.
        LGAN++ is short for LayoutGAN++.
    }
    \label{tbl:results}
    \end{minipage}
    \hfill
    \begin{minipage}{0.33\linewidth}
        \centering
        \includegraphics[width=\linewidth]{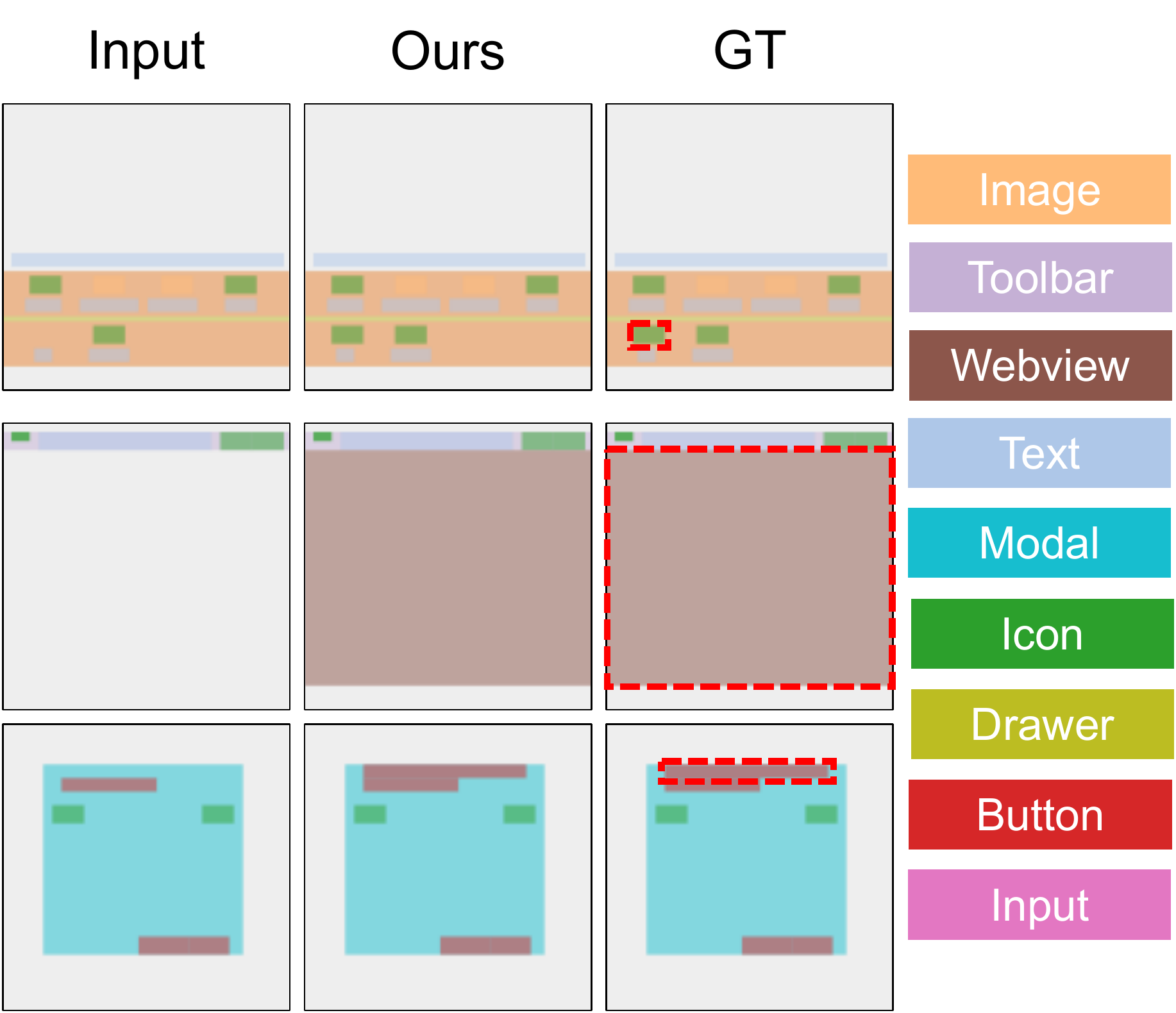}
        \captionof{figure}{
          Results in element filling using Rico dataset.
          The red dotted box indicates the target element to be predicted.
        }
        \label{fig:res_all_rico_elem}
    \end{minipage}
    \setlength{\tabcolsep}{6pt}
\end{table*}

\subsection{Training Details}
We use 256-dimensional latent representations within the encoder, Transformer blocks, and decoder.
For the Transformer blocks we use the one from DeepSVG~\cite{carlier2020deepsvg}.
We apply a dropout probability of 0.1 to all the dropout layers.
We train the model with a batch size of 256 sequences for 500 epochs in all
the experiments.
We use Adam with learning rate of 1e-4, $\beta_{1}=0.9$, $\beta_{2}=0.99$, and
L2 weight decay of 1e-2.
In experiments on Rico, we make FlexDM take positional embedding as the additional input, since otherwise the model is unable to distinguish elements having a completely similar set of attributes, which often occurs in POS prediction.

\subsection{Quantitative Evaluation}
\label{subsec:quantitative_evaluation}
We test three models based on our proposed framework to clarify the contribution of both explicit multi-task learning and pre-training.
\\ \textbf{Ours-IMP}:
As in the standard masked language modeling such as BERT~\cite{devlin-etal-2019-bert}, we randomly mask 15\% of the fields during training.
Since this randomized training is called implicit multi-task learning~\cite{radford2019language}, we call it Ours-IMP.
\\ \textbf{Ours-EXP}:
All the tasks are explicitly and jointly trained in a single model by sampling the masking patterns corresponding to each task. For simplicity, $T$ tasks introduced in \cref{subsubsec:actual_tasks} are uniformly sampled in a mini-batch.
\\ \textbf{Ours-EXP-FT}:
This is our entire model. We use weights of the model trained on IMP, and fine-tune the model. The rest of the training is the same as Ours-EXP.

We compare these models with the following baselines, some of which are adapted from existing task-specific models to our multi-task, multi-attribute, and arbitrary masking setting with minimal modification.
\\ \textbf{Expert}:
We train the network individually for each task.
Note that the number of the parameters used in this variant is $T$ times larger than our models.
\\ \textbf{Most-frequent}:
We calculate the statistics of the training dataset.
For a categorical attribute, we count the occurrences and pick the most frequent category.
For a numerical attribute, we compute the average because the numerical attributes that we use are only image and text features.
\\ \textbf{BERT~\cite{devlin-etal-2019-bert}}:
We convert all the fields into a single sequence and process them with Transformer blocks. This evaluates the effect of element-wise embedding discussed in~\cref{subsec:arch}.
\\ \textbf{BART~\cite{lewis2020bart}}:
BART employs an encoder-decoder-based sequence-to-sequence model for pre-training text generation models by masked language modeling.
We replace our Transformer blocks with the blocks from BART.
\\ \textbf{CVAE~\cite{jyothi2019layoutvae,lee2020neural}}:
Recent methods for conditional layout generation such as LayoutVAE~\cite{jyothi2019layoutvae} and NDN~\cite{lee2020neural} employ Conditional VAE~\cite{sohn2015learning} in an auto-regressive manner.
We replace our Transformer block and decoder parts with CVAE variants used in~\cite{jyothi2019layoutvae,lee2020neural} and predict the fields in an element-by-element manner.
Note that the full version of NDN contains relation prediction and layout refinement modules in addition to CVAE modules.
We omit the full NDN pipeline evaluation due to their specific approach.
\\ \textbf{CanvasVAE~\cite{yamaguchi2021canvasvae}}:
CanvasVAE is for an unconditional generation.
Although direct comparison is impossible, we adapt CanvasVAE to our setting, similar to other baselines.

\Cref{tbl:results} summarizes the performance of all the models.
Our full model (Ours-EXP-FT) is almost comparable to Expert model while being much more efficient in the number of parameters.
Ours-IMP exhibits moderate performance, resulting in a better initial weight for fine-tuning in Ours-EXP-FT.
We can see that most of the compared baselines perform clearly worse compared to Ours-EXP.
The result suggests that applying existing Transformer models for sequence modeling or conditional layout generation models is not enough in our challenging setting.
POS-prediction in Rico is the exceptional case, where most of the methods fail because of the larger number of elements compared to the benchmark setup in the literature~\cite{Kikuchi2021} (nine at maximum).

{
\begin{figure*}[t]
  \centering
  \setlength{\tabcolsep}{1pt}
  \newcommand{\fig}[1]{\includegraphics[width=0.16\textwidth]{./images/results/crello_prediction/#1.png}}
  \begin{tabular}{cccccc}
    \multicolumn{2}{c}{ATTR prediction} & \multicolumn{2}{c}{TXT prediction} & \multicolumn{2}{c}{IMG prediction} \\
    Input & Output & Input & Output & Input & Output \\
    \multirow{2}{*}{\shortstack{\fig{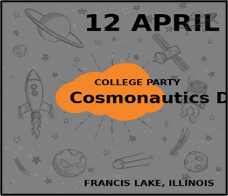} \\ \fig{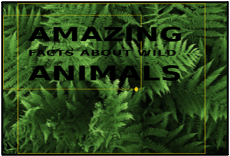}}} &
    \multirow{2}{*}{\shortstack{\fig{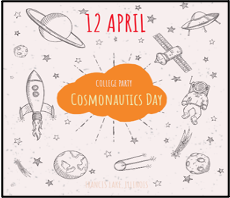} \\ \fig{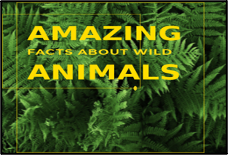}}} &
    \multirow{2}{*}{\shortstack{\fig{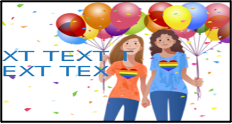} \\ \fig{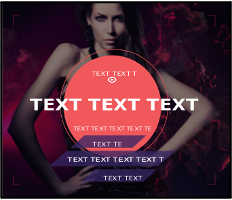}}} &
    \multirow{2}{*}{\shortstack{\fig{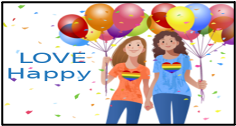} \\ \fig{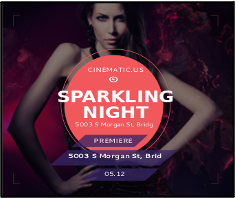}}} &
    \multirow{2}{*}{\shortstack{\fig{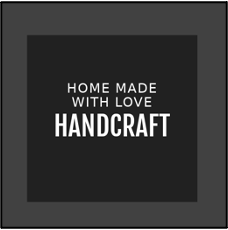} \\ \fig{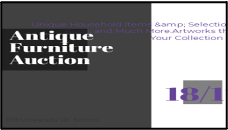}}} &
    \multirow{2}{*}{\shortstack{\fig{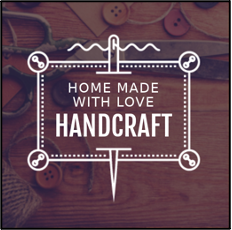} \\ \fig{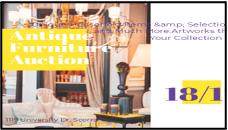}}} \\
    & & & & & \\
    \addlinespace[24ex] \multicolumn{4}{c}{POS prediction} & \multicolumn{2}{c}{Element filling} \\
    Output (bbox.) & Output (img.) & Output (bbox.) & Output (img.) & Input & Output \\
    \fig{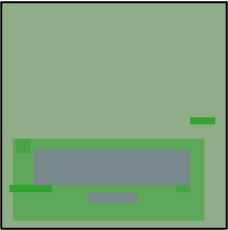} & \fig{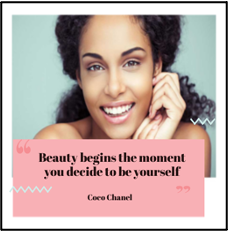} & \fig{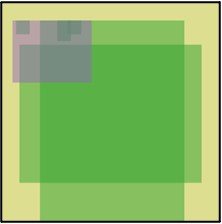} & \fig{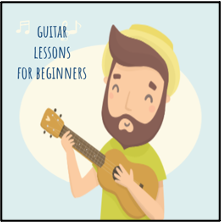} & \fig{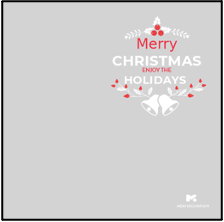} & \fig{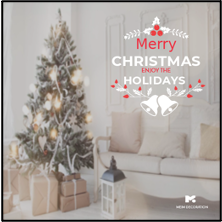} \\
  \end{tabular}

  \caption{
    Prediction of FlexDM (Ours-EXP-FT trained on Crello). FlexDM jointly handles a large variety of design tasks with a single Transformer-based model.
    In the input of ATTR/TXT/IMG prediction, the target fields assigned \texttt{[MASK]} are visualized using fixed default values (\ie, black for text color, gray for image and solid fill, `TEXT' for text).
    In POS prediction, we additionally show the layout of the elements. The correspondence between the color and type of the element is as follows: green = \textit{vector shape}, magenta = \textit{image}, purple = \textit{text}, yellow = \textit{solid fill}.
    Best viewed with zoom and color.
  }
  \label{fig:crello_prediction}
\end{figure*}
}

\subsection{Qualitative Evaluation}
We show the prediction quality of our full FlexDM (Ours-EXP-FT) for Rico dataset in the element-filling task in \cref{fig:res_all_rico_elem}.
For Rico, we show a color map indicating the position and type information.
In \cref{fig:crello_prediction}, we show the prediction of our full FlexDM (Ours-EXP-FT) on all the target design tasks.
For visualizing predicted low-dimensional image and text features, we conduct a nearest neighbor search to retrieve actual images and texts using the assets in the test subset, following CanvasVAE~\cite{yamaguchi2021canvasvae}.

\subsection{Ablation Study}
\label{subsec:ablation_study}

In this section, we perform several ablation experiments in the Crello dataset, as shown in \cref{tbl:ablation_study}.
We demonstrate that our design choices non-trivially affect the final performance of FlexDM.
\\ \noindent \textbf{Task-specific Embedding}:
The previous work~\cite{Hu_2021_ICCV} on unifying multiple tasks in a single Transformer uses small task-specific learnable query embedding to feed information of the current task explicitly.
We append the query as $h_{0}^{\text{enc}}$ at the beginning of $h^{\text{enc}} = \{h_{1}^{\text{enc}}, h_{2}^{\text{enc}}, \ldots, h_{S}^{\text{enc}}\}$ and train the model.
The result suggests the benefit of the embedding is marginal.
We conjecture that the model implicitly captures the task information from the masked inputs in our setting.
\\ \noindent \textbf{Attention}:
Here we study the importance of self-attention to model the inter-element relationship by training a model without self-attention.
We increase the number of layers to eight to roughly match the total number of parameters with Ours-EXP.
As expected, the result clearly suggests the importance of modeling the inter-element relationship.
\\ \noindent \textbf{Additional Loss}:
Our objective function in \cref{eq:objective} only considers reconstruction.
One may argue that incorporating adversarial losses such as those used in LayoutGAN++~\cite{Kikuchi2021} could improve the model.
While we tried our best in implementing and tuning the additional adversarial loss, we did not find a clear benefit in adversarial training.

\begin{table}[t]
    \setlength{\tabcolsep}{3pt}
    \centering
    \caption{
        Ablation study results in Crello dataset.
        Top two results are highlighted in \textbf{bold} and \underline{underline}, respectively.
    }
    \label{tbl:ablation_study}
    \begin{tabular}{cl ccccc} \toprule
              & Model         & ELEM           & POS               & ATTR              & IMG               & TXT               \\ \midrule
              & Ours-EXP      & \textbf{0.499} & \underline{0.218} & \textbf{0.679}    & \underline{0.948} & \underline{0.952} \\
        (i)   & w/ task-ID    & \underline{0.496}          & \textbf{0.222}    & 0.674             & \textbf{0.949}    & \textbf{0.953}    \\
        (ii)  & w/o attention & 0.446          & 0.208             & 0.605             & 0.939             & 0.947             \\
        (iii) & w/ adv.       & \textbf{0.499} & 0.215             & \underline{0.677} & \underline{0.948} & \underline{0.952} \\  %
        \bottomrule
    \end{tabular}
    \setlength{\tabcolsep}{6pt}
\end{table}

\subsection{Comparison with Task-specific Baselines}
In this section, we show that our data-driven masked field prediction model can match or even surpasses task-specific approaches.
We perform experiments in two tasks: 1) single text styling and 2) single text box placement.
Since each task uses partially overlapping set of attributes, we train our model for each single task for fair comparison.
Note that we are unable to compare to contextual images filling~\cite{wang2020learning} discussed in \cref{subsec:rw_computational_assistance} due to their task setup where they retrieve an image only from pre-defined sets used during training.

{
\begin{figure*}[t]
  \centering
  \setlength{\tabcolsep}{1pt}
    \newcommand{\length}{0.21}
    \newcommand{\smrow}[1]{
      \includegraphics[width=\length\textwidth]{./images/results/smarttext/input/#1.png} &
      \includegraphics[width=\length\textwidth]{./images/results/smarttext/smarttext/#1.png} &
      \includegraphics[width=\length\textwidth]{./images/results/smarttext/unid/#1.png} &
      \includegraphics[width=\length\textwidth]{./images/results/smarttext/gt/#1.png}
    }
    \begin{tabular}{cccc}
     Input (background) & SmartText+~\cite{li2021harmonious} & Ours & Ground truth \\
     \smrow{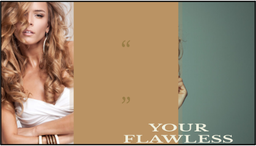} \\
     \smrow{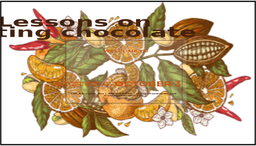} \\
     \smrow{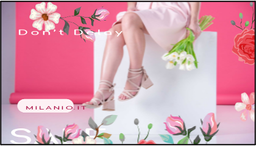} \\
    \end{tabular}
  \caption{
    Qualitative comparison of single text box placement with SmartText+~\cite{li2021harmonious}.
    Best viewed with zoom and color.
  }
  \label{fig:smarttext}
\end{figure*}
}

\subsubsection{Single Text Styling}
Zhao~\etal~\cite{zhao2018modeling} propose an MLP-based model to predict desirable font properties for a \textit{single} text box (\ie, font emb., color, and size),
given context in web designs.
We consider that each design is a document with one text and two image elements, and regard all the context information as attributes in the elements so that we can just apply FlexDM.
We implement Zhao~\etal~\cite{zhao2018modeling} with the following minor difference, since the code is not publicly available.
We quantize the color and size into 16 bins and 64 bins, respectively.
We did not apply data augmentation using the external dataset, since the dataset used for the augmentation is not available.
We show the results in \cref{tbl:results_ctx}.
The metrics are accuracy for font color and size, and cosine similarity for font type, which is represented by a low-dimensional embedding.
We can clearly see that our model is comparable to the task-specific model.
\begin{table}[t]
    \setlength{\tabcolsep}{5pt}
    \centering
    \caption{
        Comparison of models for font properties prediction in CTXFont dataset~\cite{zhao2018modeling}.
        The average and standard deviation of three runs are reported.
        The values are multiplied by 100x for visibility.
    }
    \label{tbl:results_ctx}
    \begin{tabular}{lcccccc} \toprule
        Model                   & Color                  & Size                   & Emb.                   & Avg.                   \\ \midrule
        Zhao~\etal~\cite{zhao2018modeling} & 45.8\std{2.9}          & 19.9\std{3.1}          & \textbf{79.2}\std{0.5}          & 48.2\std{1.2}          \\
        Ours                    & \textbf{54.2}\std{0.7} & \textbf{24.2}\std{0.1} & 77.7\std{1.3} & \textbf{52.0}\std{0.5} \\ \bottomrule
    \end{tabular}
    \setlength{\tabcolsep}{6pt}
\end{table}

\subsubsection{Single Text Box Placement}
\begin{table}
    \centering
    \captionof{table}{
        Quantitative evaluation of models for single text box placement in Crello dataset.
        The samples are divided into two groups: no other text box available (Single) and some text boxes available as the context (Multiple).
    }
    \label{tbl:smarttext}
    \begin{tabular}{l cc cc} \toprule
                                           & \multicolumn{2}{c}{Single} & \multicolumn{2}{c}{Multiple}                                     \\ \cmidrule{2-3} \cmidrule{4-5}
                                           & IoU~$\uparrow$             & BDE~$\downarrow$             & IoU~$\uparrow$ & BDE~$\downarrow$ \\ \midrule
        SmartText+~\cite{li2021harmonious} & 0.047                      & 0.262                        & 0.023          & 0.300            \\
        Ours                               & \textbf{0.357}             & \textbf{0.098}               & \textbf{0.110} & \textbf{0.141}   \\
        ~~w/o image                        & 0.355                      & 0.100                        & 0.103          & 0.156            \\
        ~~w/o text                         & 0.350                      & 0.106                        & 0.086          & 0.178            \\ \bottomrule
    \end{tabular}
    \setlength{\tabcolsep}{6pt}
\end{table}

Li~\etal~\cite{li2021harmonious} propose to predict the size and position of a single text box given a natural image and aspect ratio of the text box.
We perform comparison in Crello dataset, since the dataset used for their model training and evaluation is not publicly available.
We evaluate the performance in terms of the intersection over union (IoU) and boundary displacement error (BDE)~\cite{li2021harmonious}.
As shown in the upper half of \cref{tbl:smarttext}, our model clearly outperforms Li~\etal~\cite{li2021harmonious}'s model.
To measure the contribution of multi-modal features to the prediction, we exclude each of them and train the model.
The results in the lower half of \cref{tbl:smarttext} suggest that those features contribute to the better performance.
Some results are shown in \cref{fig:smarttext}.

\section{Limitation and Discussion}
As image and text generation quality is astonishingly improving, one may want to generate images and texts directly. However, retrieval-based generation is still a practical option. For instance, due to clients' requests, designers often need to use images from private collections or public photo stock services such as Adobe Stock or Shutterstock. Moreover, some people avoid using generated images or text as there are controversies about the legal and ethical issues of AI-generated images.

Our model does not support design tasks that cannot be framed as masked field prediction.
We do not consider unconditional generation; \ie, generating a complete document without input.
Extending FlexDM to an unconditional scenario requires us to apply a generative formulation instead of BERT-style masked modeling, and we leave such formulation as future work.
However, we believe that our model nicely fits in a common application scenario where there exist initial design materials to start with.

The model’s performance decreases when the input document has more elements.
Whether bigger models or datasets alleviate the issue is worth investigating.
Developing other evaluation metrics would be helpful for further analysis since current metrics simply evaluate reconstruction performance.
In conditional generation, the input context may correspond to
multiple possible outputs, especially when the input context is sparse (\eg,
label sets).
Modeling such variability as in layout generation models~\cite{jyothi2019layoutvae,Kikuchi2021,inoue2023layout} would be an exciting direction.

{\small
\bibliographystyle{ieee_fullname}
\bibliography{egbib}

\begin{thebibliography}{10}\itemsep=-1pt

\bibitem{agrawala2011design}
Maneesh Agrawala, Wilmot Li, and Floraine Berthouzoz.
\newblock Design principles for visual communication.
\newblock {\em Communications of the ACM}, 54(4), 2011.

\bibitem{argyriou2006multi}
Andreas Argyriou, Theodoros Evgeniou, and Massimiliano Pontil.
\newblock Multi-task feature learning.
\newblock In {\em NeurIPS}, 2006.

\bibitem{arroyo2021variational}
Diego~Martin Arroyo, Janis Postels, and Federico Tombari.
\newblock Variational transformer networks for layout generation.
\newblock In {\em CVPR}, 2021.

\bibitem{chongyang2021uibert}
Chongyang Bai, Xiaoxue Zang, Ying Xu, Srinivas Sunkara, Abhinav Rastogi,
  Jindong Chen, and Blaise Agüera~y Arcas.
\newblock {UIBert}: Learning generic multimodal representations for ui
  understanding.
\newblock In {\em IJCAI}, 2021.

\bibitem{carlier2020deepsvg}
Alexandre Carlier, Martin Danelljan, Alexandre Alahi, and Radu Timofte.
\newblock {DeepSVG}: A hierarchical generative network for vector graphics
  animation.
\newblock In {\em NeurIPS}, 2020.

\bibitem{caruana1997multitask}
Rich Caruana.
\newblock Multitask learning.
\newblock {\em Machine learning}, 28(1):41--75, 1997.

\bibitem{deka2017rico}
Biplab Deka, Zifeng Huang, Chad Franzen, Joshua Hibschman, Daniel Afergan, Yang
  Li, Jeffrey Nichols, and Ranjitha Kumar.
\newblock {Rico}: A mobile app dataset for building data-driven design
  applications.
\newblock In {\em UIST}, 2017.

\bibitem{svg}
Patrick Dengler, Erik Dahlstr{\"{o}}m, Doug Schepers, Jon Ferraiolo, Anthony
  Grasso, Chris Lilley, Dean Jackson, Jonathan Watt, Cameron McCormack, and Jun
  Fujisawa.
\newblock Scalable vector graphics ({SVG}) {1.1} (second edition).
\newblock {W3C} recommendation, W3C, Aug. 2011.
\newblock https://www.w3.org/TR/2011/REC-SVG11-20110816/.

\bibitem{devlin-etal-2019-bert}
Jacob Devlin, Ming-Wei Chang, Kenton Lee, and Kristina Toutanova.
\newblock {BERT}: Pre-training of deep bidirectional transformers for language
  understanding.
\newblock In {\em NAACL}, 2019.

\bibitem{evgeniou2004regularized}
Theodoros Evgeniou and Massimiliano Pontil.
\newblock Regularized multi--task learning.
\newblock In {\em KDD}, 2004.

\bibitem{fu2022doc2ppt}
Tsu-Jui Fu, William~Yang Wang, Daniel McDuff, and Yale Song.
\newblock Doc2ppt: Automatic presentation slides generation from scientific
  documents.
\newblock In {\em AAAI}, 2022.

\bibitem{guo2021vinci}
Shunan Guo, Zhuochen Jin, Fuling Sun, Jingwen Li, Zhaorui Li, Yang Shi, and Nan
  Cao.
\newblock Vinci: an intelligent graphic design system for generating
  advertising posters.
\newblock In {\em CHI}, 2021.

\bibitem{gupta2021layout}
Kamal Gupta, Alessandro Achille, Justin Lazarow, Larry Davis, Vijay Mahadevan,
  and Abhinav Shrivastava.
\newblock {LayoutTransformer}: Layout generation and completion with
  self-attention.
\newblock In {\em ICCV}, 2021.

\bibitem{ha2017neural}
David Ha and Douglas Eck.
\newblock A neural representation of sketch drawings.
\newblock In {\em ICLR}, 2018.

\bibitem{he2022masked}
Kaiming He, Xinlei Chen, Saining Xie, Yanghao Li, Piotr Doll{\'a}r, and Ross
  Girshick.
\newblock Masked autoencoders are scalable vision learners.
\newblock In {\em CVPR}, 2022.

\bibitem{he2021actionbert}
Zecheng He, Srinivas Sunkara, Xiaoxue Zang, Ying Xu, Lijuan Liu, Nevan Wichers,
  Gabriel Schubiner, Ruby Lee, Jindong Chen, and Blaise~Aguera y Arcas.
\newblock {ActionBert}: Leveraging user actions for semantic understanding of
  user interfaces.
\newblock In {\em AAAI}, 2021.

\bibitem{Hu_2021_ICCV}
Ronghang Hu and Amanpreet Singh.
\newblock {UniT}: Multimodal multitask learning with a unified transformer.
\newblock In {\em ICCV}, 2021.

\bibitem{inoue2023layout}
Naoto Inoue, Kotaro Kikuchi, Edgar Simo-Serra, Mayu Otani, and Kota Yamaguchi.
\newblock {LayoutDM: Discrete Diffusion Model for Controllable Layout
  Generation}.
\newblock In {\em CVPR}, 2023.

\bibitem{jaegle2022perceiverio}
Andrew Jaegle, Sebastian Borgeaud, Jean-Baptiste Alayrac, Carl Doersch, Catalin
  Ionescu, David Ding, Skanda Koppula, Daniel Zoran, Andrew Brock, Evan
  Shelhamer, et~al.
\newblock Perceiver {IO}: A general architecture for structured inputs \&
  outputs.
\newblock In {\em ICLR}, 2022.

\bibitem{jaegle2021perceiver}
Andrew Jaegle, Felix Gimeno, Andy Brock, Oriol Vinyals, Andrew Zisserman, and
  Joao Carreira.
\newblock Perceiver: General perception with iterative attention.
\newblock In {\em ICML}, 2021.

\bibitem{jyothi2019layoutvae}
Akash~Abdu Jyothi, Thibaut Durand, Jiawei He, Leonid Sigal, and Greg Mori.
\newblock {LayoutVAE}: Stochastic scene layout generation from a label set.
\newblock In {\em CVPR}, 2019.

\bibitem{kikuchi2023generative}
Kotaro Kikuchi, Naoto Inoue, Mayu Otani, Edgar Simo-Serra, and Kota Yamaguchi.
\newblock Generative colorization of structured mobile web pages.
\newblock In {\em WACV}, 2023.

\bibitem{kikuchi2021modeling}
Kotaro Kikuchi, Mayu Otani, Kota Yamaguchi, and Edgar Simo-Serra.
\newblock Modeling visual containment for web page layout optimization.
\newblock {\em Computer Graphics Forum}, 40(7), 2021.

\bibitem{Kikuchi2021}
Kotaro Kikuchi, Edgar Simo-Serra, Mayu Otani, and Kota Yamaguchi.
\newblock Constrained graphic layout generation via latent optimization.
\newblock In {\em ACM MM}, 2021.

\bibitem{kokkinos2017ubernet}
Iasonas Kokkinos.
\newblock {Ubernet}: Training a universal convolutional neural network for
  low-, mid-, and high-level vision using diverse datasets and limited memory.
\newblock In {\em CVPR}, 2017.

\bibitem{kong2022blt}
Xiang Kong, Lu Jiang, Huiwen Chang, Han Zhang, Yuan Hao, Haifeng Gong, and
  Irfan Essa.
\newblock {BLT}: bidirectional layout transformer for controllable layout
  generation.
\newblock In {\em ECCV}, 2022.

\bibitem{lee2020neural}
Hsin-Ying Lee, Weilong Yang, Lu Jiang, Madison Le, Irfan Essa, Haifeng Gong,
  and Ming-Hsuan Yang.
\newblock Neural design network: Graphic layout generation with constraints.
\newblock In {\em ECCV}, 2020.

\bibitem{lewis2020bart}
Mike Lewis, Yinhan Liu, Naman Goyal, Marjan Ghazvininejad, Abdelrahman Mohamed,
  Omer Levy, Ves Stoyanov, and Luke Zettlemoyer.
\newblock {BART}: Denoising sequence-to-sequence pre-training for natural
  language generation, translation, and comprehension.
\newblock In {\em ACL}, 2020.

\bibitem{li2021harmonious}
Chenhui Li, Peiying Zhang, and Changbo Wang.
\newblock Harmonious textual layout generation over natural images via deep
  aesthetics learning.
\newblock {\em IEEE TMM}, 2021.

\bibitem{li2019layoutgan}
Jianan Li, Jimei Yang, Aaron Hertzmann, Jianming Zhang, and Tingfa Xu.
\newblock {LayoutGAN}: Generating graphic layouts with wireframe
  discriminators.
\newblock In {\em ICLR}, 2019.

\bibitem{li2020attribute}
Jianan Li, Jimei Yang, Jianming Zhang, Chang Liu, Christina Wang, and Tingfa
  Xu.
\newblock Attribute-conditioned layout gan for automatic graphic design.
\newblock {\em IEEE TVCG}, 2020.

\bibitem{li2021selfdoc}
Peizhao Li, Jiuxiang Gu, Jason Kuen, Vlad~I Morariu, Handong Zhao, Rajiv Jain,
  Varun Manjunatha, and Hongfu Liu.
\newblock {SelfDoc}: Self-supervised document representation learning.
\newblock In {\em CVPR}, 2021.

\bibitem{liu2019end}
Shikun Liu, Edward Johns, and Andrew~J Davison.
\newblock End-to-end multi-task learning with attention.
\newblock In {\em CVPR}, 2019.

\bibitem{lok2001survey}
Simon Lok and Steven Feiner.
\newblock A survey of automated layout techniques for information
  presentations.
\newblock In {\em SmartGraphics}, 2001.

\bibitem{lopes2019learned}
Raphael~Gontijo Lopes, David Ha, Douglas Eck, and Jonathon Shlens.
\newblock A learned representation for scalable vector graphics.
\newblock In {\em CVPR}, 2019.

\bibitem{lu2023unifiedio}
Jiasen Lu, Christopher Clark, Rowan Zellers, Roozbeh Mottaghi, and Aniruddha
  Kembhavi.
\newblock Unified-{IO}: A unified model for vision, language, and multi-modal
  tasks.
\newblock In {\em ICLR}, 2023.

\bibitem{lu202012}
Jiasen Lu, Vedanuj Goswami, Marcus Rohrbach, Devi Parikh, and Stefan Lee.
\newblock 12-in-1: Multi-task vision and language representation learning.
\newblock In {\em CVPR}, 2020.

\bibitem{odonovan2015designscape}
Peter O'Donovan, Aseem Agarwala, and Aaron Hertzmann.
\newblock {DesignScape}: Design with interactive layout suggestions.
\newblock In {\em CHI}, 2015.

\bibitem{odonovan2014learning}
Peter O’Donovan, Aseem Agarwala, and Aaron Hertzmann.
\newblock Learning layouts for single-pagegraphic designs.
\newblock {\em IEEE TVCG}, 20(8), 2014.

\bibitem{qiu2023color}
Qianru Qiu, Xueting Wang, Mayu Otani, and Yuki Iwazaki.
\newblock Color recommendation for vector graphic documents based on
  multi-palette representation.
\newblock In {\em WACV}, 2023.

\bibitem{radford2021learning}
Alec Radford, Jong~Wook Kim, Chris Hallacy, Aditya Ramesh, Gabriel Goh,
  Sandhini Agarwal, Girish Sastry, Amanda Askell, Pamela Mishkin, Jack Clark,
  et~al.
\newblock Learning transferable visual models from natural language
  supervision.
\newblock In {\em ICML}, 2021.

\bibitem{radford2019language}
Alec Radford, Jeffrey Wu, Rewon Child, David Luan, Dario Amodei, Ilya
  Sutskever, et~al.
\newblock Language models are unsupervised multitask learners.
\newblock {\em OpenAI blog}, 1(8):9, 2019.

\bibitem{rebuffi2017learning}
Sylvestre-Alvise Rebuffi, Hakan Bilen, and Andrea Vedaldi.
\newblock Learning multiple visual domains with residual adapters.
\newblock In {\em NeurIPS}, 2017.

\bibitem{rebuffi2018efficient}
Sylvestre-Alvise Rebuffi, Hakan Bilen, and Andrea Vedaldi.
\newblock Efficient parametrization of multi-domain deep neural networks.
\newblock In {\em CVPR}, 2018.

\bibitem{sohn2015learning}
Kihyuk Sohn, Honglak Lee, and Xinchen Yan.
\newblock Learning structured output representation using deep conditional
  generative models.
\newblock 2015.

\bibitem{vaswani2017attention}
Ashish Vaswani, Noam Shazeer, Niki Parmar, Jakob Uszkoreit, Llion Jones,
  Aidan~N Gomez, {\L}ukasz Kaiser, and Illia Polosukhin.
\newblock Attention is all you need.
\newblock In {\em NeurIPS}, 2017.

\bibitem{wang2020learning}
Guolong Wang, Zheng Qin, Junchi Yan, and Liu Jiang.
\newblock Learning to select elements for graphic design.
\newblock In {\em ICMR}, 2020.

\bibitem{wang2022ofa}
Peng Wang, An Yang, Rui Men, Junyang Lin, Shuai Bai, Zhikang Li, Jianxin Ma,
  Chang Zhou, Jingren Zhou, and Hongxia Yang.
\newblock Ofa: Unifying architectures, tasks, and modalities through a simple
  sequence-to-sequence learning framework.
\newblock In {\em ICML}, 2022.

\bibitem{xie2021canvasemb}
Yuxi Xie, Danqing Huang, Jinpeng Wang, and Chin-Yew Lin.
\newblock {CanvasEmb}: Learning layout representation with large-scale
  pre-training for graphic design.
\newblock In {\em ACM MM}, 2021.

\bibitem{xu2020layoutlm}
Yiheng Xu, Minghao Li, Lei Cui, Shaohan Huang, Furu Wei, and Ming Zhou.
\newblock {LayoutLM}: Pre-training of text and layout for document image
  understanding.
\newblock In {\em KDD}, 2020.

\bibitem{xu2020layoutlmv2}
Yang Xu, Yiheng Xu, Tengchao Lv, Lei Cui, Furu Wei, Guoxin Wang, Yijuan Lu,
  Dinei Florencio, Cha Zhang, Wanxiang Che, et~al.
\newblock {LayoutLMv2}: Multi-modal pre-training for visually-rich document
  understanding.
\newblock In {\em ACL}, 2020.

\bibitem{yamaguchi2021canvasvae}
Kota Yamaguchi.
\newblock {CanvasVAE}: Learning to generate vector graphics documents.
\newblock In {\em ICCV}, 2021.

\bibitem{yang2016automatic}
Xuyong Yang, Tao Mei, Ying-Qing Xu, Yong Rui, and Shipeng Li.
\newblock Automatic generation of visual-textual presentation layout.
\newblock {\em TOMM}, 12(2), 2016.

\bibitem{yuan2021infocolorizer}
Lin-Ping Yuan, Ziqi Zhou, Jian Zhao, Yiqiu Guo, Fan Du, and Huamin Qu.
\newblock Infocolorizer: Interactive recommendation of color palettes for
  infographics.
\newblock {\em IEEE TVCG}, 28(12), 2021.

\bibitem{zhang2014facial}
Zhanpeng Zhang, Ping Luo, Chen~Change Loy, and Xiaoou Tang.
\newblock Facial landmark detection by deep multi-task learning.
\newblock In {\em ECCV}, 2014.

\bibitem{zhao2018modeling}
Nanxuan Zhao, Ying Cao, and Rynson~WH Lau.
\newblock Modeling fonts in context: Font prediction on web designs.
\newblock {\em Comput. Graph. Forum}, 37(7), 2018.

\bibitem{zheng2019content}
Xinru Zheng, Xiaotian Qiao, Ying Cao, and Rynson~WH Lau.
\newblock Content-aware generative modeling of graphic design layouts.
\newblock {\em ACM TOG}, 38(4), 2019.

\end{thebibliography}
}

\end{document}